\documentclass[sigplan,screen,authorversion,nonacm]{acmart} 

\usepackage{times}
\usepackage{url}
\usepackage{listings, lstautogobble}
\usepackage{xspace}

\usepackage[nameinlink]{cleveref}
\usepackage{hyperref}

\usepackage{balance}

\newcommand{\eg}{\hbox{\emph{e.g.}}\xspace}

\newcommand{\etc}{\hbox{\emph{etc.}}\xspace}
\newcommand{\vs}{\hbox{\emph{vs.}}\xspace}

\newcommand{\cf}[1]{\texttt{#1}}

\newboolean{showcomments}
\setboolean{showcomments}{true} 
\ifthenelse{\boolean{showcomments}}{

}{
}

\newcommand{\hitrate}{55\%\xspace}
\newcommand{\latencyredux}{$5\times$\xspace}
\newcommand{\cogsredux}{$56\%$\xspace}

\setcopyright{cc}
\setcctype{by}
\copyrightyear{2026}
\acmYear{2026}


\begin{document}

\title{On Device Agentic Operation Caches}
\subtitle{Classifier-Centric NL-to-Action Generation}

\author{Moghis Fereidouni}
\affiliation{%
  \institution{University of Kentucky}
  \city{}
  \country{}
}
\email{moghis.fereidouni@uky.edu}

\author{Anthony Arnold}
\affiliation{%
  \institution{University of Kentucky}
  \city{}
  \country{}
}
\email{anthony.arnold@uky.edu}

\author{Sumit Gulwani}
\affiliation{%
  \institution{Microsoft}
  \city{}
  \country{}
}
\email{sumitg@microsoft.com}

\author{Mark Marron}
\affiliation{%
  \institution{University of Kentucky}
  \city{}
  \country{}
}
\email{mark.marron@uky.edu}

\author{A.B. Siddique}
\affiliation{%
  \institution{University of Kentucky}
  \city{}
  \country{}
}
\email{ab.siddique@uky.edu}

\begin{abstract}
Agentic AI is increasingly being embedded in software applications to provide natural language 
interfaces to features and functionality. In most cases these agents are powered by enterprise 
($100+$ billion parameter) or frontier class large language models that require substantial computational resources run 
and depend on cloud hosted inference to handle the task of transforming natural language inputs into 
actionable software operations. This reliance on cloud-hosted inference introduces substantial network 
latency on top of LLM inference times, creates data privacy concerns, and, given the costs of running these 
models, can rapidly escalate expenses associated with supporting agentic features. 

This paper introduces a novel means of converting the NL-to-Action problem from a generative one into 
a classification-centric formulation via \emph{on-device operation caches}.
These caches allow an agentic system to handle frequently occurring classes of actions completely on-device -- reducing latency, 
enhancing privacy, and lowering operational costs. We show that for a classic NL-to-Formula task, generating Excel 
Formula in response to user requests, this approach reduces total inference cost by \cogsredux when compared to cloud-only 
model-routing based inference and, on cache hits, reduces the latency to response latency by \latencyredux.
\end{abstract}

\maketitle

\section{Introduction}
\label{sec:introduction}

Adding natural language interfaces to software applications has, driven by the capabilities of large language models, 
become increasingly common. The most direct approach to providing natural language interfaces is through the use of 
agentic AI, powered by enterprise ($100+$ billion parameter) or frontier class large language models that are run via 
cloud-hosted inference. While effective, this approach can rapidly escalate cost-of-goods (COGS) and the combination 
of network latency and time-to-first token can lead to poor user experience.

A first general step in addressing these cost challenges is the introduction of model-routing mechanisms that can 
intelligently select a suitable model for a query based on its estimated complexity and resource requirements. However, 
this does little to improve the latency issue, and as the routing decisions are designed to error on the side of 
selecting more capable models the costs remain high. 

However, when when we look at the overall workflow for handling natural language queries in a specific application, such 
as spreadsheets, it is quickly apparent that queries follow a pareto style distribution, and typically involve a 
relatively small set of recurring operations, suggesting the potential for developing a classifier to identify these 
recurring operations and enable \emph{semantic operation caching} to improve efficiency.

Taking inspiration from traditional caching mechanisms, which are proven to be a highly effective strategy for 
improving performance and reducing costs on workloads with these features, this paper proposes the introduction of an on-device operation cache that can store and reuse frequently occurring operations.
This \emph{on-device semantic operation cache} is used to handle the most frequently used operations locally -- entirely eliminating the need for repeated cloud interactions 
and thus eliminate latency, privacy, and cost concerns for these common operations. 


A critical difference from simple domains and standard cache implementation, where the elements in the cache are constant values and matching is simply equality, 
is that in our setting the cache elements are parameterized operation idioms and the context and NL inputs will not exactly match these existing symbolic cache values. 
Thus the key insight in our approach is the use of a semantic matcher that can classify parameterized operation idioms based on context and natural language input.
This \emph{semantic matcher}, will be used to perform a combination of classification (cache lookup) to identify the most relevant cached 
operation idiom for a given NL input and table context. It will then extract and synthesize the final operation or formula by filling in the parameters of the matched operation 
idiom based on the specific context and NL input.

This combination of an on-device operation cache and a semantic matcher allows for efficient handling of recurring operations, significantly reducing latency, cost, and privacy 
concerns, while passing complex or unusual operations to the cloud when necessary, thus maintaining high accuracy in interpreting natural language queries. 
The contributions of this paper are as follows: 
\begin{itemize}
\item Using the insight that, for major classes of common tasks, ``generation is nearly classification'' and can be efficiently approximated by classification + light synthesis, we introduce \emph{on-device semantic operation caches} for efficiently handling frequently occurring application operations.
\item We present a mechanism for constructing a semantic matcher that can be run on consumer hardware, provides an economically impactful hit (coverage) rate, and do so without compromising result accuracy (\Cref{sec:cache}).
\item We provide a fully automated method for constructing and maintaining the on-device semantic operation cache based entirely on observed usage patterns (\Cref{sec:construction} and \Cref{sec:extraction}).
\item We demonstrate the effectiveness of the on-device semantic operation cache in the high-value domain of generating Excel spreadsheet formulas from natural language queries (\Cref{sec:evaluation})
-- achieving a hit rate of approximately \hitrate while maintaining a (correctness) rate of over $96\%$, which translates into a \cogsredux reduction in inference costs and a \latencyredux speedup 
in query response time (when cache hits occur).
\end{itemize}

\section{Operation Caches}
\label{sec:cache}

This section provides an overview of how the proposed on-device operation caches interact with agentic natural language integration in a software application. 
For the purposes of this paper we focus on Microsoft Excel spreadsheets as an exemplar of this type of interface as, given the ubiquity, power, and semi-technical 
abilities of the user-base, it is a prime use case for agent augmented natural language interactions.

\subsection{Agentic Natural Language Integration}
\label{sec:nlinterface}

\begin{figure}
    \centering
    \includegraphics[width=\linewidth]{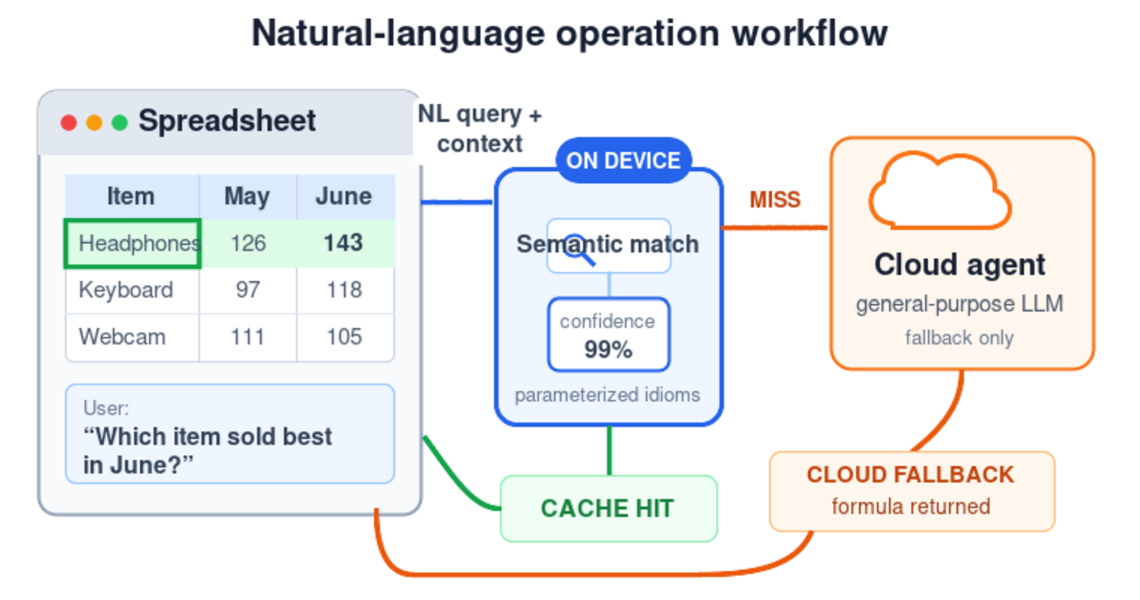}
    \caption{Interaction between on-device operation caches and agentic natural language integration.}
    \Description{A spreadsheet sends a natural-language query and table context to an on-device semantic cache. A high-confidence cache hit returns a specialized formula locally; a miss routes the request to a general-purpose cloud agent.}
    \label{fig:workflow}
\end{figure}

\Cref{fig:workflow} illustrates the interaction between the on-device operation caches and agentic natural language support in a spreadsheet. 
In the standard workflow with the spreadsheet a user can ask a question or issue a command in natural language. This NL input, along with meta-data about 
the sheet context, is sent to a cloud based agent for processing and response generation. Once the response is generated, it is sent back to the spreadsheet application, 
which then updates the user interface accordingly.

In our example this natural language might be the command ``Which item sold best in June'' within the context of a sales spreadsheet that has the months as columns and items, along 
with sales figures, as the entries. Modern LLM-powered agents have little trouble with a query of this nature, and in practice, respond with an (Excel) formula that when 
run on the users sheet will select the corresponding row -- \cf{INDEX(Sales[Item], MATCH(MAX(Sales[July]), Sales[July], 0))}. 

\subsection{On-Device Operation Cache Mechanism}
\label{sec:on_device_cache}

\begin{figure}
    \centering
    \includegraphics[width=\linewidth]{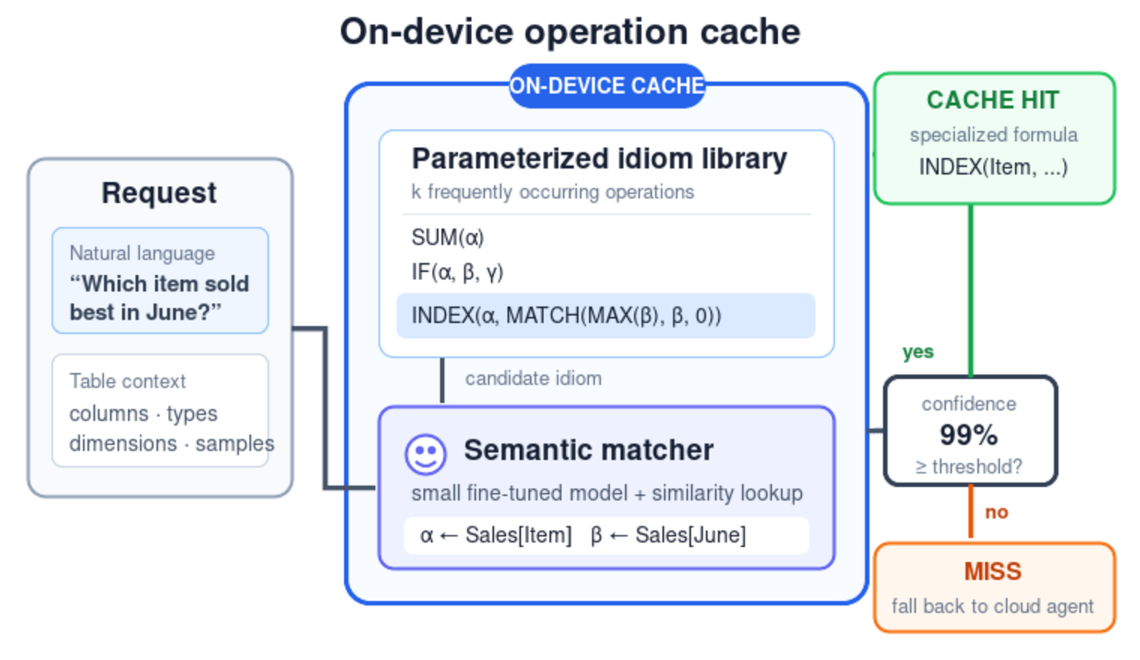}
    \caption{On-device operation cache architecture.}
    \Description{Natural-language input and table context enter an on-device cache containing parameterized operation idioms and a semantic matcher. The matcher binds formula parameters and applies a confidence threshold, returning either a specialized formula or a miss that triggers cloud fallback.}
    \label{fig:cache}
\end{figure}

The detailed architecture of the cache mechanism is shown in \Cref{fig:cache}. The on-device cache acts as an intermediary 
between the user interface and the cloud-based AI model processing layer. Given a set of the $k$-most frequently occurring operation idioms~\cite{nlyze,miningidioms,idioms} the cache will take a users NL input along with any 
table metadata, such as column names, row counts, data types, and sample data elements, and from these either generate the appropriate specialized formula (based on the cache) or report a miss. 
In the case of a cache miss, the system will fall back to cloud-based processing to handle the operation, ensuring that all user requests can be fulfilled even if they are not part of the frequently 
occurring set.

The critical difference from simple domains, where the elements in the cache are constant values and matching is defined as exact value equality, is that in our setting the cache elements 
are parameterized operation idioms and the context and NL inputs will not exactly match symbolic cache values. Thus, we require a more sophisticated matching mechanism, shown as the 
\emph{semantic matcher}, in the cache architecture. This matcher will be used to perform a combination of lookup and similarity-based matching to identify the most relevant cached operation 
idiom for a given NL input and table context. It will also be used to synthesize the final operation or formula by filling in the parameters of the matched operation idiom based on the 
specific context and NL input.

The semantic matcher implementation utilizes a small neural network~\cite{gemma3} (or System One~\cite{jev}) model that is fine-tuned on the specific $k$-most frequently occurring operation idioms over a set of user interactions. 
Using a combination of model prediction and internal confidence-scoring we classify each user input as a cache hit or miss, determining whether to use the cached operation or fall back to cloud-based processing. 

In this scenario the cost of a false negative, erroneously reporting a cache miss, is an increase in costs while the cost of a false positive, incorrectly using the cached operation, could lead to 
incorrect results being presented to the user. Adjusting these parameters allows us to balance the trade-off between cache hit rate and the risk of incorrect results, optimizing the overall 
performance and reliability. As described in \Cref{sec:evaluation}, for our Excel spreadsheet scenario, we are able to maintain a precision (correctness) rate of over $96\%$, which matches 
or exceeds the performance of cloud-based processing alone, while achieving a hit rate of approximately $55\%$.

In our example scenario, the user input ``Which item sold best in June'' will trigger a match with the cached operation idiom corresponding to -- \cf{INDEX($\alpha$, MATCH(MAX($\beta$), $\beta$, 0))}. The 
neural network will then select the appropriate values to replace the parameters in the idiom, $\alpha \rightarrow \cf{Sales[Item]}$ and $\beta \rightarrow \cf{Sales[June]}$, and subsequently evaluate 
the confidence score for this match (internally generating a $99\%$ score). Given that this confidence score exceeds the predefined threshold, the system will proceed to use the cached operation, and 
return the desired formula with all operations occurring locally and with an end-to-end latency of ${\sim}0.3s$ (as opposed to ${\sim}1.6s$ for cloud-based processing).

In the case of a more complex or unusual user query, say ``Build an ARR model for growth in sales based on the values for Q2'', the system will not identify any suitable cached operation. In 
such cases, it will simply return a special \cf{MISS} indicator and the system will fall back to cloud-based processing to handle the request.

\section{Automatic Cache Construction}
\label{sec:construction}

Cache construction involves training a model to predict the appropriate operations for given inputs, based on previously observed interactions. As our goal is to have a fully automated 
pipeline we design this process to process a stream of raw user interaction data and, from that, to automatically construct the operation cache.

\subsection{Cache Construction Input Data}
Formally, the input to the cache construction process is a dataset of user interactions, where each interaction consists of a user query and the relevant context plus the corresponding 
operation that should be executed by the application. This dataset can be represented as data tuples:
\[
\mathcal{D} = \{(q_i, c_i, o_i)\}_{i=1}^N
\]
where \(q_i\) is the user query, \(c_i\) is the relevant context, and \(o_i\) is the operation to be executed.

In our systems $q_i$ is a natural language query, $c_i$ is a JSON representation of the table headers and size, cell types, and sample rows, and $o_i$ is an Excel Formula. In our 
running example we would see data in the format shown in \autoref{fig:data_format}.

\begin{figure*}
\caption{Example of the input data format for cache construction. Each entry consists of a natural language query, the relevant spreadsheet context, and the corresponding spreadsheet formula.}
\label{fig:data_format}
\centering
\scriptsize
\begin{tabular}{|l|l|l|}
\hline
Query & Context & Operation \\
\hline
\text{"Sum the values in column A"} & \{"headers": ["A", "B"], \dots]\} & \text{"SUM(A1:A10)"} \\
\text{"Total of Sales"} & \{"name": "Table1", "headers": ["Items", "Quantity", "Sales"], \dots]\} & \text{"SUM(Table1[Sales])"} \\
\text{"how mnay wins"} & \{"name": "Games", "headers": ["Opponent", "Win", "Loss"], \dots]\} & \text{"SUM(Games[Win])"} \\
\ldots & \ldots & \ldots \\
\text{"Which item sold best in June"} & \{"name": "Sales", "headers": ["Item", "Jan", \ldots ], \dots]\} & \text{"INDEX(Sales[Item], MATCH(MAX(Sales[July]), Sales[July], 0))"} \\
\hline
\end{tabular}
\end{figure*}

From this input data stream we first process the queries to extract the highest frequency idioms. In theory we could mine arbitrary idioms~\cite{miningidioms} but for the purposes of this 
work we rely on the simpler approach of mechanically replacing all variables/constants in the formula with $\alpha$-placeholders and treating the resulting pattern as an idiom. 
The systems then clusters the NL/context inputs based on $\alpha$-equivalence of their corresponding idioms, grouping together inputs that map to the same abstracted formula 
pattern, and ranks them from largest to smallest cluster size. We refer to this process as $\alpha$-clustering.

The resulting clustered dataset is:
\[
\mathcal{D}_{\alpha} = \{(\{(q_i, c_i) \mid \alpha\text{-rename}(o_i) \equiv f_k^{\alpha}\}, f_k^{\alpha})\}_{k=1}^K
\]
where $\alpha\text{-rename}(o_i)$ denotes the alpha-renamed version of \(o_i\) and $f_k^{\alpha}$ is the corresponding idiom.

Applying this to the concrete example in \Cref{fig:data_format} would produce the idiom abstracted and clustered dataset shown in \Cref{fig:clustered_data}.

\begin{figure*}
\caption{Idiom abstracted and clustered dataset corresponding to the input data format shown in \Cref{fig:data_format}.}
\label{fig:clustered_data}
\centering
\scriptsize
\begin{tabular}{|l|l|}
\hline
Cluster & $\alpha$-Operation \\
\hline
$\begin{aligned} 
                  \left\{ \begin{array}{ll} 
                    \text{"Sum the values in column A"}  & \{"headers": ["A", "B"], \dots]\} \\
                    \text{"Total of Sales"} & \{"name": "Table1", "headers": ["Items", "Quantity", "Sales"], \dots]\} \\
                    \text{"how mnay wins"} & \{"name": "Games", "headers": ["Opponent", "Win", "Loss"], \dots]\} \\
                    \ldots & \ldots \\
                  \end{array} \right\} 
                \end{aligned}$ & \text{"SUM($\alpha$)"} \\
\ldots & \ldots \\
$\begin{aligned} 
                  \left\{ \begin{array}{ll} 
                    \text{"Which item sold best in June"} & \{"name": "Sales", "headers": ["Item", "Jan", \ldots ], \dots]\} \\
                    \ldots & \ldots \\
                  \end{array} \right\} 
                \end{aligned}$ & \text{"INDEX($\alpha$, MATCH(MAX($\beta$), $\beta$, 0))"} \\
\hline
\end{tabular}
\end{figure*}

The data in \Cref{fig:clustered_data} shows the results of the idiom extraction and clustering process. All of the NL inputs (even on different table context) that match the same abstracted formula 
pattern are grouped together into the same cluster. In this case the three summation operations, even though the language and contexts differ, are identified as mapping to the same abstracted formula 
$\text{"SUM($\alpha$)"}$. This idiomatic abstraction allows the system to recognize patterns across different contexts, facilitating more efficient caching and retrieval of operations. 
Similarly, the operation of finding the best-selling item in a specific month is abstracted to the formula $\text{"INDEX($\alpha$, MATCH(MAX($\beta$), $\beta$, 0))"}$, and might be grouped in with 
other variants, \eg finding the highest scoring player in a tournament, would also be recognized as instances of the same idiomatic pattern.

\subsection{Cache Construction Training}
After the clustering process we are ready to begin training the semantic mapper. At this point we need to make a key parametric decision in the system design. Specifically selecting the target cache-hit 
rate. This decision will influence the trade-off between the effectiveness of the cache and the eventual ability to ensure high precision. In the training we must select the number of idioms $F_i$ to 
be used as labels in the cache. As each label corresponds to a specific idiomatic pattern cluster, we can select more labels until we reach a desired hit-rate target. Once the target cache-hit rate is 
determined, we can proceed to select the appropriate number of idioms $F_i$ to include in the cache and perform a fine-tuning of the semantic mapper to optimize cache performance.

Model fine-tuning supervision uses \emph{learned refusal} which trains on all examples, mapping included bin examples to the appropriate target program with all $\alpha$ values replaced with concrete 
values, and mapping the remaining queries to the literal token \cf{MISS}. 

In our example the model will be trained to map queries corresponding to the summation idiom to the complete formula, \eg ``sum the values in column A'' $\rightarrow$ $\text{SUM(A2:A10)}$, and 
queries corresponding other unselected clusters, such as the ARR examples, will be mapped to the literal token \cf{MISS}.

\subsection{Confidence-Gated Routing}
\label{sec:routing}

At inference time the cache model decodes greedily. For a generated sequence $\hat{y} = (\hat{y}_1, \ldots, \hat{y}_T)$ we compute a sequence-level
confidence as the exponentiated, length-normalized log-likelihood over content tokens (excluding padding and end-of-sequence):
\begin{equation}
c(x, \hat{y}) \;=\;
\exp\!\Bigl( \frac{1}{|\mathcal{T}|} \sum_{t \in \mathcal{T}}
\log p_\theta(\hat{y}_t \mid x, \hat{y}_{<t}) \Bigr),
\label{eq:confidence}
\end{equation}
i.e., the geometric mean of per-token probabilities, so that scores are comparable across generations of different lengths. 

The final cache-hit decision is then based on both 1) the model's refusal gate and 2) the confidence gate. Specifically, a cache hit occurs only if the model 
produces a formula \emph{and} the confidence exceeds the threshold; otherwise, the query is routed to the fallback model.
\begin{equation}
R(x) \;=\;
\begin{cases}
\hat{y} & \text{if } \hat{y} \neq \cf{MISS} \;\wedge\; c(x,\hat{y}) \ge \tau,\\[2pt]
F(x) & \text{otherwise.}
\end{cases}
\label{eq:router}
\end{equation}
The refusal gate captures queries the model \emph{knows} are out-of-head; the confidence gate additionally catches in-head queries where the particular
generation is unreliable. As both signals are byproducts of a single greedy decode this design avoids extra model calls and allows us to perform the cache 
lookup and generation in one pass.

\section{Evaluation}
\label{sec:evaluation}

To evaluate the effectiveness of our caching approach, we conduct a series of experiments measuring both cache performance and model robustness under various conditions on a 
NL2Formula~\cite{zhao2024nl2formula} benchmark. This dataset is comprised of $70,799$ paired NL queries and their corresponding formulas, associated with $21,670$ tables, 
and covering a wide range of domains and query types.

In all experiments our cache is configured with a target hit rate of $60\%$ and a target of at least $95\%$ precision. We regard this precision as the minimum acceptable 
level for reliable cache performance and, given the inherent ambiguity or noise in a large dataset like NL2Formula, is near the upper bound of what can be reasonably 
expected without overfitting.

\subsection{Cache Model Training}

The default cache model we select is a fully fine-tuned small instruction-tuned decoder (Gemma~3~1B~\cite{gemma3}) which was selected for a balance between ability to handle 
the semantic nuance needed to handle complex domains like NL queries on arbitrary tables while still being small and efficient enough to easily run on consumer hardware.
Fine-tuning is performed with the Unsloth stack~\cite{unsloth}. 
In all of our evaluations, we use this default cache model unless otherwise specified.

For the fine-tuning process, we use a learning rate of $2\times10^{-5}$ and an effective batch size of $32$ (a per-device batch size of $8$ with $4$ gradient-accumulation steps). 
We optimize with 8-bit AdamW with a weight decay of $0.001$ and a linear learning-rate schedule with $5$ warm-up steps. We use a maximum sequence length of $1024$ tokens and compute 
the loss only on the response tokens. The model is trained for a maximum of $10$ epochs, with early stopping based on validation loss to prevent overfitting. Validation loss 
is evaluated every $300$ steps with a patience of $3$ evaluations, and the checkpoint with the lowest validation loss is kept. The training data for this process is a special 
hold-out subset of the NL2Formula dataset which is not used in any other part of the experiments. All finetune runs were performed on a single NVIDIA GeForce RTX 3090 GPU 
and took a maximum of $6$ hours.

\subsection{Cache Performance}
The primary metrics we evaluate are the ability of the cache to correctly serve queries, measured in terms of hit rate and precision, and secondarily, the overall impact this has on 
latency and inference costs. 

To evaluate these features, we compare a baseline Excel agent where every request is handled in the cloud using GPT-5, with model routing enabled, against our proposed approach where a local 
cache with Gemma~3~1B is used first, and only queries with low confidence are passed to the cloud. 

\begin{table}
    \caption{Cache performance comparison between using a local cache with cloud fall back and using cloud only. Hit rate indicates the fraction of queries served by the local cache. Cost 
    is the total inference cost (in USD at API costs) to run all queries in the experiment. Avg. Latency indicates the average time taken to serve a query, with separate values for cache hits and misses 
    -- cloud only has a single value as all queries are served from the cloud.}
    \label{tab:cache_performance}
    \centering
    \begin{tabular}{lcc}
        \toprule
        & Cached + Cloud & Cloud Only \\
        \midrule
        Hit Rate & 55\% & - \\
        Cost (USD) & \$0.67 & \$1.54 \\
        Correctness & 97\%(hit)/83\%(miss) & 85\% \\
        Avg. Latency (s) & 0.31(hit)/1.6(miss) & 1.5 \\
        \bottomrule
    \end{tabular}
\end{table}

The results in \Cref{tab:cache_performance} demonstrate that using a local cache with cloud fall back significantly reduces both latency and inference costs while maintaining a high hit rate. As 
shown in the \emph{Hit Rate} row, the local cache is able to serve a substantial fraction of queries, while still maintaining over a $95\%$ correctness rate, thus reducing the number of requests 
that are processed remotely. This directly translates into substantial reductions in inference costs, reducing them by \cogsredux, and on cache hits, bringing the time to serve a query down to just 
0.31 seconds. 

Interestingly the average latency for cache misses is larger than the average over all queries in the cloud-only setting. This is a result of the cache handling the majority of simple queries, 
leaving complex, and inference time-consuming queries, to be processed by the cloud. Additionally, the use of a very selective 
and high-correctness cache actually increases the overall precision of the system. The cloud only implementation 
achieves a correctness rate of $85\%$ \vs the combined system rate of $91\%$!

\subsection{Local Model Variants}
Based on the excellent performance of the cache with cloud fall back, a critical research question is if local models 
alone can achieve similar benefits. To address this, we evaluated the performance of various local only standard model 
variants with the same prompting strategy used for the cloud inference with model sizes from $1B$ to $27B$. 

\begin{table}
    \caption{Performance of local Gemma~3 instruction-tuned model variants on the NL2Formula benchmark. The correctness percentage 
    is over all queries (full recall).}
    \label{tab:local_model_variants}
    \centering
    \begin{tabular}{lc}
        \toprule
        Model & Correct Percentage \\
        \midrule
        Gemma~3--1B & $1\%$ \\
        Gemma~3--4B & $6\%$ \\
        Gemma~3--12B & $17\%$ \\
        Gemma~3--27B & OOM \\
        \bottomrule
    \end{tabular}
\end{table}

As \Cref{tab:local_model_variants} shows, the untuned local models are almost completely ineffective on tasks with 
the complexity of generating Excel formulas. And even on a workstation class machine, with an NVIDIA 5070 GPU (16GB VRAM), 
the larger local model runs out of memory (OOM). This matches the results in~\cite{promptintern} which reports that 
large cloud models struggle with the complexity of generating Excel formulas and that even aggressive finetunes of 
local models cannot provide sufficient precision for this task -- with GTP4o giving only $66\%$ correctness rate and 
a full SFT finetune of Qwen3-8B reaches an $80\%$ correctness rate.

\subsection{Feature Ablation and Analysis}
To understand the impact of the various cache features on performance, we conduct feature ablation studies on the two critical aspects of the system -- the impact of the internal 
confidence gate and the impact of noise on the cache performance.

\begin{figure}
    \centering
    \includegraphics[width=\linewidth]{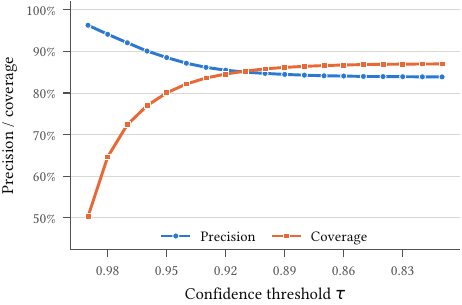}
    \caption{Impact of confidence gating on cache performance. Hit rate and correctness curves are shown for varying values of the confidence threshold.}
    \label{fig:confidence_gating}
\end{figure}

\paragraph{Confidence Gating}
To assess the impact of confidence gating on cache performance, we compare the correctness and entropy of predictions at various hit rates. As shown in \Cref{fig:confidence_gating}, 
the entropy based internal confidence score generated from the cache model is able to effectively gate low-confidence results and fall back to the cloud when necessary. Setting the 
confidence threshold to a high value (over $0.99$) is able to keep the correctness high, above $95\%$, while still achieving a hit rate of $50.2\%$. As we lower the threshold the 
hit-rate increases, to a max of $87\%$, while correctness gradually decreases down to $84\%$.

\paragraph{Noise and Robustness}
To evaluate the robustness of the cache to noisy inputs, we perturb the natural-language queries with keyboard-typo noise using the \cf{nlpaug} library~\cite{ma2019nlpaug}. For each query, $30\%$ of the 
words are selected, and within each selected word, $10\%$ of the characters are replaced by a neighboring key on a QWERTY keyboard. Words that must match the table's contents, such as string literals 
and numeric constants referenced by the ground-truth formula, are protected from perturbation so that each noisy query still has a well-defined answer.

\begin{figure}
    \centering
    \includegraphics[width=\linewidth]{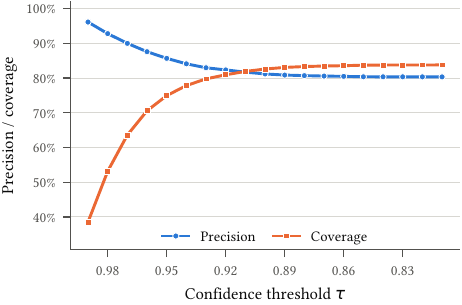}
    \caption{Impact of noise on cache performance. Hit rate and correctness curves are shown for varying levels of noise in the natural language queries.}
    \label{fig:noise_robustness}
\end{figure}

The results in \Cref{fig:noise_robustness} show that the cache's performance is robust to additional noise in the natural language queries. As shown in the figure, the cache is able to maintain 
the desired high correctness levels (staying at $96\%$) despite the presence of noise while experiencing a drop in the hit-rate by $12$pp to $38.6\%$. This demonstrates that, even with the small model sizes, the 
cache is able to handle noise and other perturbations in the inputs effectively.

\section{Data Collection and Specialization}
\label{sec:extraction}

A key consideration in the design of the cache mechanism, and training setup, presented in this work is enable automation of both an initial training corpus \emph{and} 
online (incremental) retraining of the cache based on new user interactions. In addition to these baseline capabilities, the system is also designed to support specialization 
for individual teams or organizations by allowing the cache to adapt to particular usage patterns, vocabularies, operation sets, and preferences.

\paragraph{Automatic Data Corpus Construction}
As described in \Cref{sec:construction}, the required input data for training the cache are triples of the form $(q_i, c_i, o_i)$, where $q_i$ represents the user query, 
$c_i$ represents the context or conditions under which the query is made, and $o_i$ represents the corresponding operation or output generated by the system. As such 
these training values are exactly what the system inputs and, cloud based agent, outputs are during normal (cache unassisted operation). Thus, the initial training corpus 
can be directly constructed from these interactions, ensuring that the cache is trained on realistic and representative examples from the outset.

Further as shown in the evaluations (\Cref{sec:evaluation}), an performant cache can be constructed from a relatively modest initial training corpus, $55k$ examples for the 
NL2Formula dataset, and then trained using an entry-level ML workstation, highlighting the efficiency of the proposed approach.

\paragraph{Specialization}
Beyond simply producing a general-purpose cache, the system is designed to support specialization for individual teams or organizations. This allows the cache to adapt to particular 
usage patterns, vocabularies, operation sets, and preferences, thereby improving its effectiveness and relevance in specific contexts. Given the ability to automatically build 
a training dataset and quickly re-build the cache model, it becomes feasible to maintain a specialized and up-to-date cache that continuously adapts to the evolving needs and 
behaviors of specific teams or organizations.

In this scenario once an initial cache is built and deployed, say to teams in finance, HR, \etc we can monitor the use of each cache and gather data on hit/miss rates along the way.
From this dynamic information it becomes possible to rebuild caches specialized to the different use patterns and vocabularies used by individual 
teams, and given the hardware requirements for training, for many organizations to maintain and update specialized caches entirely on-premises.

\section{Related Work}

\paragraph{Natural language to Formula:} The dream of providing natural language interfaces to applications, and spreadsheets in particular, has been a long-standing topic of research. 
Early work on semantic parsing for natural language to formula translation focused on using statistical machine translation techniques~\cite{semanticparse} and structural program 
synthesis approaches~\cite{nlyze}. The advent of large language models provided a new means for approaching the problem and has been explored in recent 
works~\cite{zhao2024nl2formula,hermes,nl2formulafewshot} that demonstrate the effectiveness of large language models in translating natural language to formulas on spreadsheets.

\paragraph{Routing and Optimization:} The cost of running a frontier scale LLM for every user query has been studied from a number of directions. The most general is the development 
of routing~\cite{ong2024routellm} and caching strategies that aim to minimize redundant computations~\cite{chen2023frugalgpt,bang2023gptcache} and optimize resource usage. These general  
techniques are designed to work accross a wide range of (perhaps dynamically varying) sets of LLM promts.

\paragraph{Specialized Models and Local Caching:} Recent work has investigated the possibility of local models and caching strategies to reduce the reliance on large-scale LLMs for 
every query. This includes approaches such as \cite{bang2023gptcache,meancache,vcache}, which leverage local caching mechanisms based on semantic embbeddings and similarity search, 
or fine-tuning based on a known set of prompts~\cite{promptintern}. 

The two most closely related approache is the work of Chakraborty et.al.~\cite{gencache} which, similar to the work in this paper, explore the use of local caching mechanisms
that map from natural language queries to a template for generating the corresponding result. In~\cite{gencache}, the system generates a program that, when run, will compute the literal 
result corresponding to the query and associates it with a regular expression pattern. When a new query matches the regex, the cached program is executed to produce the result. 
As a result this system is highly effective for simple datasets where the natural language variability and presence of noise is limited. 
    
In contrast, our approach generalizes observed patterns in result formula, to create a set of reusable templates that can be applied to new queries, and associates these with a 
semantic embbedding. This enables the system, described in this paper, to effectively match new natural language queries in complex domains and with a rich variety NL formulations 
including paraphrased queries, synonyms, typos, and variations in sentence structure.

\section{Conclusion}
Using a novel means of converting the NL-to-Action problem from a generative one into 
a classification-centric formulation via \emph{on-device operation caches}. These caches allow 
an agentic system to handle frequently occurring actions completely on-device. 
In our evaluation we showed that, for the task of converting natural language into Excel 
formulas, this approach is effective in reducing latency (by \latencyredux for cache hits), enhancing privacy, 
lowering operational costs (by \cogsredux), and perhaps counter intuitively increasing the overall accuracy of 
the results provided by the system (increasing precision by $6$pp). As this cache can be built entirely from 
observing existing system behavior we believe it is a practical and effective approach to improving the 
overall performance of agentic systems.


\begin{acks}
    This work was supported in part by a gift from Microsoft.
\end{acks}

\balance
\bibliographystyle{ACM-Reference-Format}
\bibliography{bibfile}

@inproceedings{zhao2024nl2formula,
  author    = {Wei Zhao and Zhitao Hou and Siyuan Wu and Yan Gao and Haoyu Dong and Yao Wan and Hongyu Zhang and Yulei Sui and Haidong Zhang},
  title     = {{NL2Formula: Generating Spreadsheet Formulas from Natural Language Queries}},
  booktitle = {Findings of the Association for Computational Linguistics: EACL 2024},
  year      = {2024},
  publisher = {Association for Computational Linguistics}
}

@inproceedings{bang2023gptcache,
  author    = {Fu Bang},
  title     = {{GPTCache: An Open-Source Semantic Cache for {LLM} Applications Enabling Faster Answers and Cost Savings}},
  booktitle = {Proceedings of the 3rd Workshop for Natural Language Processing Open Source Software (NLP-OSS)},
  year      = {2023},
  publisher = {Association for Computational Linguistics}
}

@article{chen2023frugalgpt,
  author  = {Lingjiao Chen and Matei Zaharia and James Zou},
  title   = {{FrugalGPT: How to Use Large Language Models While Reducing Cost and Improving Performance}},
  journal = {arXiv preprint arXiv:2305.05176},
  year    = {2023}
}

@article{ong2024routellm,
  author  = {Isaac Ong and Amjad Almahairi and Vincent Wu and Wei-Lin Chiang and Tianhao Wu and Joseph E. Gonzalez and M. Waleed Kadous and Ion Stoica},
  title   = {{RouteLLM: Learning to Route {LLMs} with Preference Data}},
  journal = {arXiv preprint arXiv:2406.18665},
  year    = {2024}
}

@article{gemma3,
  author  = {{Gemma Team}},
  title   = {{Gemma 3 Technical Report}},
  journal = {arXiv preprint arXiv:2503.19786},
  year    = {2025}
}

@misc{ma2019nlpaug,
  title={{NLP Augmentation}},
  author={Edward Ma},
  howpublished={https://github.com/makcedward/nlpaug},
  year={2019}
}

@misc{jev,
  title={{Introducing System One Models and JEV}},
  author    = {{Typesafe AI}},
  howpublished={\url{https://typesafe.ai/blog/introducing-system-one-models-and-jev}},
  year={2026}
}

@misc{unsloth,
  author       = {Daniel Han and Michael Han},
  title        = {{Unsloth: Finetune LLMs Faster}},
  howpublished = {\url{https://github.com/unslothai/unsloth}},
  year         = {2024}
}

@ARTICLE{idioms,
  author={Allamanis, Miltiadis and Barr, Earl T. and Bird, Christian and Devanbu, Premkumar and Marron, Mark and Sutton, Charles},
  journal={IEEE Transactions on Software Engineering}, 
  title={{Mining Semantic Loop Idioms}}, 
  year={2018},
  volume={44},
  number={7},
  pages={651-668},
}

@inproceedings{miningidioms,
author = {Allamanis, Miltiadis and Sutton, Charles},
title = {{Mining idioms from source code}},
year = {2014},
publisher = {Association for Computing Machinery},
booktitle = {Proceedings of the 22nd ACM SIGSOFT International Symposium on Foundations of Software Engineering (FSE)}
}

@inproceedings{nlyze,
author = {Gulwani, Sumit and Marron, Mark},
title = {{NLyze: interactive programming by natural language for spreadsheet data analysis and manipulation}},
year = {2014},
publisher = {Association for Computing Machinery},
booktitle = {Proceedings of the 2014 ACM SIGMOD International Conference on Management of Data (SIGMOD)}
}

@article{gencache,
      title={{Generative Caching for Structurally Similar Prompts and Responses}}, 
      author={Sarthak Chakraborty and Suman Nath and Xuchao Zhang and Chetan Bansal and Indranil Gupta},
      year={2025},
      journal = {arXiv preprint arXiv:2511.17565}
}

@misc{vcache,
      title={{vCache: Verified Semantic Prompt Caching}}, 
      author={Luis Gaspar Schroeder and Aditya Desai and Alejandro Cuadron and Kyle Chu and Shu Liu and Mark Zhao and Stephan Krusche and Alfons Kemper and Matei Zaharia and Joseph E. Gonzalez},
      year={2026},
      journal = {arXiv preprint arXiv:2502.03771}
}

@inproceedings{meancache,
   title={{MeanCache: User-Centric Semantic Caching for LLM Web Services}},
   author={Gill, Waris and Elidrisi, Mohamed and Kalapatapu, Pallavi and Ahmed, Ammar and Anwar, Ali and Gulzar, Muhammad Ali},
   year={2025},
   booktitle={2025 IEEE International Parallel and Distributed Processing Symposium (IPDPS)}
}

@inproceedings{hermes,
    title = {{HermEs: Interactive Spreadsheet Formula Prediction via Hierarchical Formulet Expansion}},
    author = "He, Wanrong  and Dong, Haoyu  and Gao, Yihuai  and Fan, Zhichao  and Guo, Xingzhuo  and Hou, Zhitao  and Lv, Xiao  and Jia, Ran  and Han, Shi  and Zhang, Dongmei",
    booktitle = "Proceedings of the 61st Annual Meeting of the {A}ssociation for {C}omputational {L}inguistics (Volume 1: Long Papers)",
    year = "2023"
}

@inproceedings{semanticparse,
author = {Wong, Yuk Wah and Mooney, Raymond J.},
title = {{Learning for Semantic Parsing with Statistical Machine Translation}},
year = {2006},
publisher = {Association for Computational Linguistics (HLT-NAACL)}
}

@inproceedings{promptintern,
    title = {{Learning from Near-Misses: Error-Aware Contrastive Few-Shot Learning for {NL}2{F}ormula}},
    author = "Shuai, Zhihao  and Chen, Yiyun  and Ma, Maolin  and Chen, Yutong  and Qiu, Hanjia  and Xu, Jing  and Chen, Ziye  and Yang, Weikai",
    booktitle = "Proceedings of the 64th Annual Meeting of the {A}ssociation for {C}omputational {L}inguistics (Volume 1: Long Papers)",
    year = "2026"
}

@inproceedings{nl2formulafewshot,
    title = {{Learning from Near-Misses: Error-Aware Contrastive Few-Shot Learning for NL2Formula}},
    author = "Shuai, Zhihao  and Chen, Yiyun  and Ma, Maolin  and Chen, Yutong  and Qiu, Hanjia  and Xu, Jing  and Chen, Ziye  and Yang, Weikai",
    booktitle = "Proceedings of the 64th Annual Meeting of the {A}ssociation for {C}omputational {L}inguistics (Volume 1: Long Papers)",
    year = "2026"
}


\end{document}